\documentclass{article}
\usepackage{PRIMEarxiv}
\usepackage{lineno,hyperref}
\usepackage{multirow}
\usepackage[cmex10]{amsmath}
\usepackage{bm}
\usepackage{flexisym}
\usepackage{times}
\usepackage{epsfig}
\usepackage{graphicx}
\usepackage{caption}
\usepackage{amsmath}
\usepackage{amssymb}
\usepackage{caption}    
\graphicspath{{media/}}     

\title{US-GAN: On the importance of Ultimate Skip Connection for Facial Expression Synthesis
}

\author{
  Arbish Akram, Nazar Khan\\
  Department of Computer Science \\
  University of the Punjab \\
  Lahore, Pakistan\\
  \texttt{\{arbishakram, nazarkhan\}@pucit.edu.pk} \\
}

\begin{document}
\maketitle

\begin{abstract}We demonstrate the benefit of using an ultimate skip (US) connection for facial expression synthesis using generative adversarial networks (GAN). A direct connection transfers identity, facial, and color details from input to output while suppressing artifacts. The intermediate layers can therefore focus on expression generation only. This leads to a light-weight US-GAN model comprised of encoding layers, a single residual block, decoding layers, and an ultimate skip connection from input to output. US-GAN has $3\times$ fewer parameters than state-of-the-art models and is trained on $2$ orders of magnitude smaller dataset. It yields $7\%$ increase in face verification score (FVS) and $27\%$ decrease in average content distance (ACD). Based on a randomized user-study, US-GAN outperforms the state of the art by $25\%$ in face realism, $43\%$ in expression quality, and $58\%$ in identity preservation.
\end{abstract}

\keywords{Facial Expression synthesis,Generative Adversarial Network, Skip connection, Image-to-image translation, Residual block.}

\section{Introduction}\label{sec:intro}

Facial expression synthesis is an image-to-image translation task, which aims to change the expression of a given image to a desired one. Photorealistic facial expression synthesis can be useful for data augmentation to train face and expression recognition models. It is also useful for producing animations for human-computer interaction and entertainment. 
Image synthesis has received considerable attention with the arrival of generative adversarial networks (GANs) \cite{goodfellow-2014} and its conditional variant \cite{mirza2014conditional}.A GAN consists of two networks, a generator and a discriminator. It employs an adversarial learning scheme to train both networks. The generator tries to fool the discriminator by producing realistic-looking fake images while the discriminator learns to distinguish real images from fake ones. GANs have been utilized to solve many image related problems including image synthesis \cite{perarnau-2016}, image-to-image translation \cite{isola-2016, zhu-2017}, image super-resolution \cite{chen2020learning, chen2018fsrnet}, and facial attribute editing \cite{yi2017dualgan, wu2019relgan, gao2021high}. Recent GAN-based methods have shown impressive results on facial expression synthesis tasks \cite{choi-2017, pumarola2018ganimation, tang2021eggan} as well. 
Despite producing photorealistic and plausible results, these models have two limitations. First, they require large datasets to synthesize photorealistic expressions. When trained on smaller datasets, \cite{choi-2017, khan2020masked, akram2021-pixel_fes}, and \cite{chen2020domain} have shown that existing models introduce color degradation and noticeable artifacts in the synthesized expressions. 
Second, the majority of existing methods \cite{choi-2017, pumarola2018ganimation, d2021ganmut, wu2019relgan} share a common architecture \cite{johnson2016perceptual} in their generator which incurs a significantly large computational cost and prohibits deployment on resource-constrained devices.
 
\begin{figure}[t]
    \centering
    \includegraphics[width=1\linewidth]{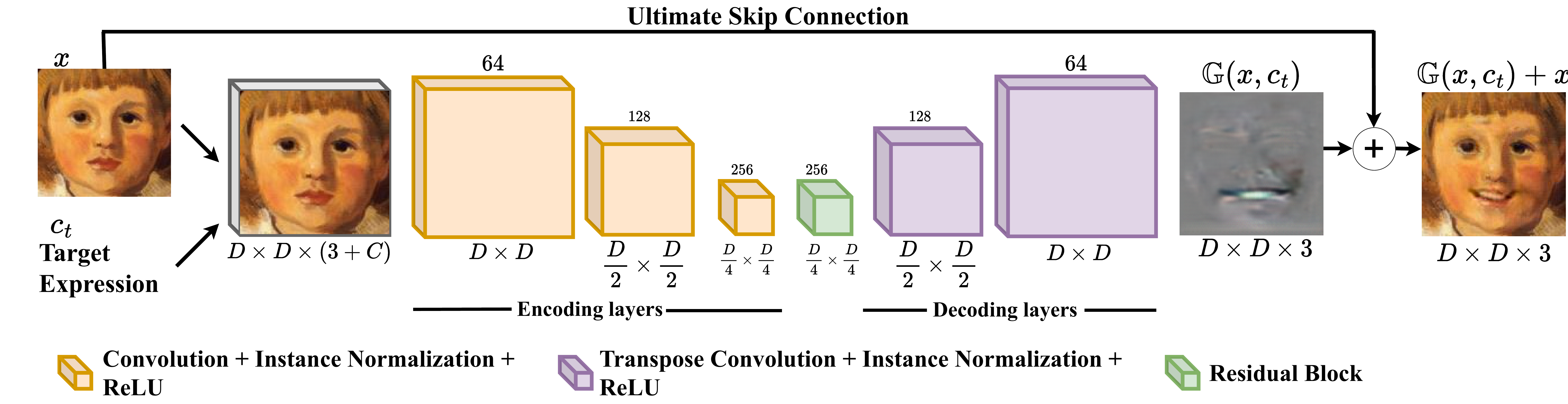}
    \caption{The generator of the proposed framework. US-GAN takes a face image with any expression and the target expression label as input and produces the output image with the target expression. It consists of four modules: i) encoding layers, ii) single residual block, iii) decoding layers, and iv) an ultimate skip connection.}
    \label{fig:architecture}
\end{figure} 
 
In this paper, we aim to build a simpler, smaller, and effective facial expression synthesis model called Ultimate Skip Generative Adversarial Network (US-GAN). The generator of the proposed US-GAN consists of four modules: encoding layers, single residual block, decoding layers and an ultimate skip connection as illustrated in Figure \ref{fig:architecture}. The encoding layers enable the network to encode key facial details. The residual block helps to refine these details. The decoding layers then decode high-level facial details from this latent encoding. Finally, we propose to directly link the input image to an output image with a skip connection, called the ultimate skip connection. We hypothesize that the inclusion of an ultimate skip connection in the generator of a GAN will lead to improved transfer of identity, facial details, and color details from input to output. Since the generator will be relieved from the task of producing such details, it will utilize it's parameters to learn expression mappings only and not suffer from color degradation and block artifacts.

The contributions of this work can be summarized as follows:
\begin{enumerate}
    \item Incorporation of an ultimate skip connection improves preservation of identity, facial details, and color details while inducing convincing expressions.

    \item The ultimate skip connection leads to a model with three times fewer parameters that can be trained using two orders of magnitude smaller dataset than the state-of-the-art GANimation model.
    
    \item US-GAN qualitatively outperforms the state of the art in realism, mapped expression and identity preservation by $25\%, 43\%$, and $58\%$ respectively.
    
    \item Quantitatively, US-GAN improves face verification score (FVS) by $7\%$ and reduces average content distance (ACD) by $27\%$ compared to the state of the art.
    
    \item US-GAN generalizes well on out-of-dataset facial images. 
    
\end{enumerate}

\begin{figure}[t]
    \centering
    \includegraphics[width=\linewidth]{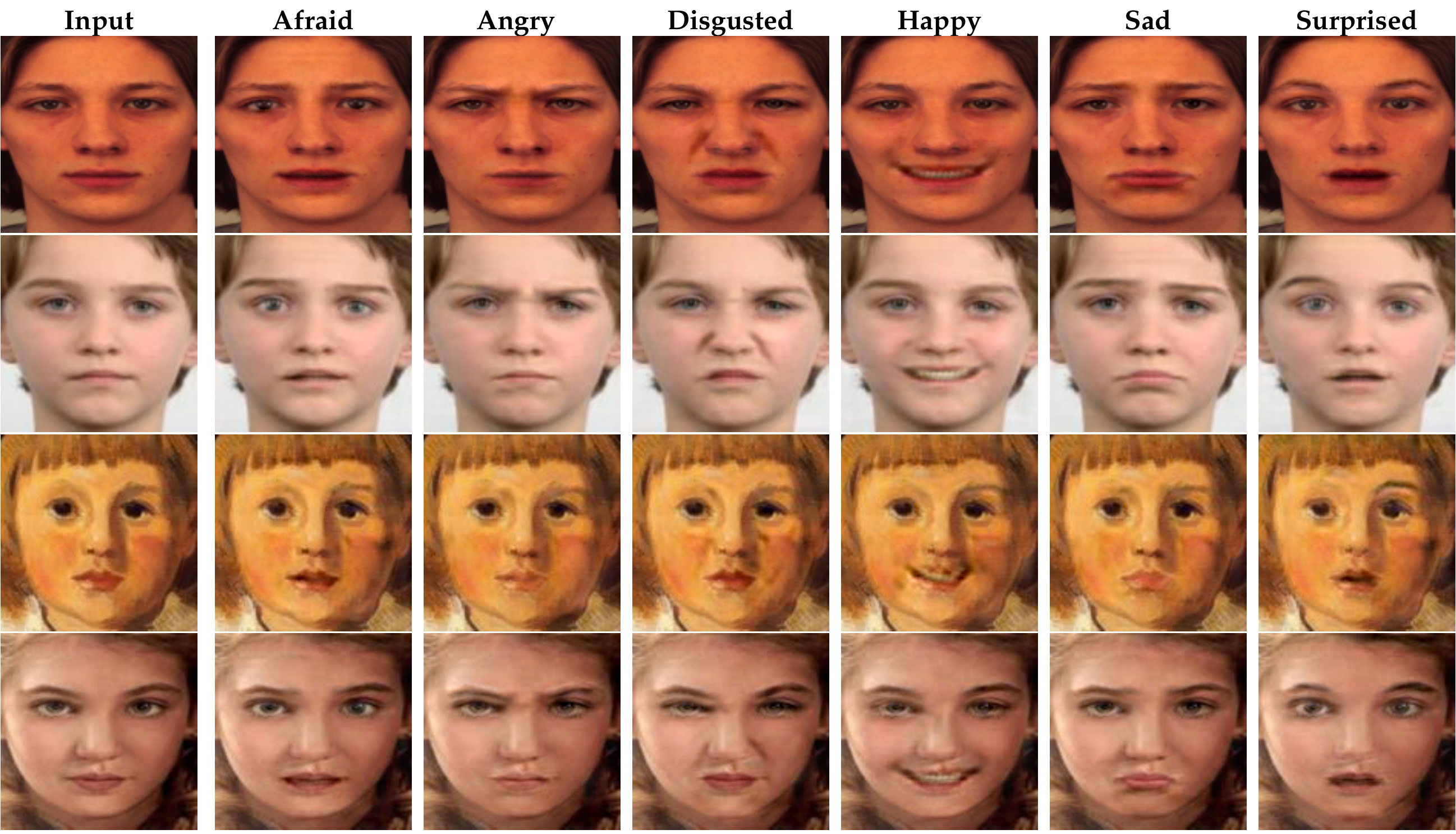}
      \caption{Facial expression synthesis results on in-dataset (top two rows) and out-of-dataset (last two rows) testing images. The proposed method trained on RaFD, KDEF and CFEE datasets generates plausible expressions while preserving identity and retaining facial details. Despite being trained on a smaller dataset, these results demonstrate both in-dataset and out-of-dataset generalization strength of the proposed method.}
    \label{fig:main_figure}
\end{figure}

The rest of the paper is structured as follows. Section \ref{sec:related_work} reviews state-of-the-art image-to-image translation and facial expression synthesis models. The proposed US-GAN architecture is explained in Section \ref{sec:material_and_methods} along with loss functions, datasets, and evaluation metrics. 
Qualitative as well as quantitative experimental results are presented in Section \ref{sec:results}. Section \ref{sec:discussion} discusses the effectiveness of our proposed method and we conclude in Section \ref{sec:conclusion}.

\section{Related Work}
\label{sec:related_work}

\textbf{Image-to-image translation} 
Generative Adversarial Network (GAN) \cite{goodfellow-2014} and its conditional variant (cGAN) \cite{mirza2014conditional} have been successfully utilized for image synthesis and image-to-image translation problems. 
The Pix2pix model \cite{isola-2016} employed a conditional GAN with $\ell_1$ image reconstruction loss to enforce generated samples to be close to target images. In order to learn the mapping between two domains using paired datasets, they utilized U-Net \cite{ronneberger-2015} and PatchGAN \cite{isola-2016} in the generator and discriminator, respectively.
Zhu et al. \cite{zhu-2017} introduced a cycle consistency loss to perform cross-domain mapping using an unpaired dataset. Their network, named CycleGAN, contains two generators and two discriminators to learn the cyclical, cross-domain mappings. They adopted Johnson et al's architecture \cite{johnson-2016perceptual} in their generator. While Pix2pix and CycleGAN can be used to learn the mappings between facial expressions, these networks fail to produce realistic expressions as demonstrated in \cite{khan2020masked} where the smaller size of facial expression synthesis datasets is suggested as a possible reason for their failure.

\textbf{Facial expression synthesis}
GANs have been widely used for facial expression synthesis. Ding et al. \cite{ding2018exprgan} proposed ExprGAN to synthesize facial expressions with controllable intensity. However, it fails to preserve the identity details of the input image. The GC-GAN model \cite{qiao2018} induces a desired facial expression on an input image. To learn mappings among multiple expressions, StarGAN \cite{choi-2017} learns mappings among multiple expressions using a single, shared generator. When trained on a small dataset, the authors reported color degradation artifacts. The GANimation model \cite{pumarola2018ganimation} trained on the large EmotioNet \cite{fabian2016emotionet} dataset extracts action units from a target face and transfers them to an input face. 
The quality of expressions is highly dependent on the extracted action units. 
Liu et al \cite{liu2019stgan} proposed an encoder-decoder architecture, called STGAN, with symmetric skip connections and selective transfer units for facial attribute manipulation. The Cascade-EF GAN \cite{wu2020cascade} generates sharper and realistic images by employing local and global attention. It focuses in a progressive manner. For out-of-dataset images, it was fine-tuned on the large AffectNet \cite{mollahosseini2017affectnet} dataset. In summary, current state-of-the-art, GAN-based, facial expression synthesis models require larger datasets for inducing satisfactory expressions on in- and out-of-dataset images. Quality significantly degrades when trained on smaller datasets. In contrast, our proposed US-GAN method produces realistic expressions on both in- and out-of-dataset images by employing only hundreds of images for training.

\section{Materials and Methods}
\label{sec:material_and_methods}

\begin{figure}[t]
    \centering
    \includegraphics[width=1\linewidth]{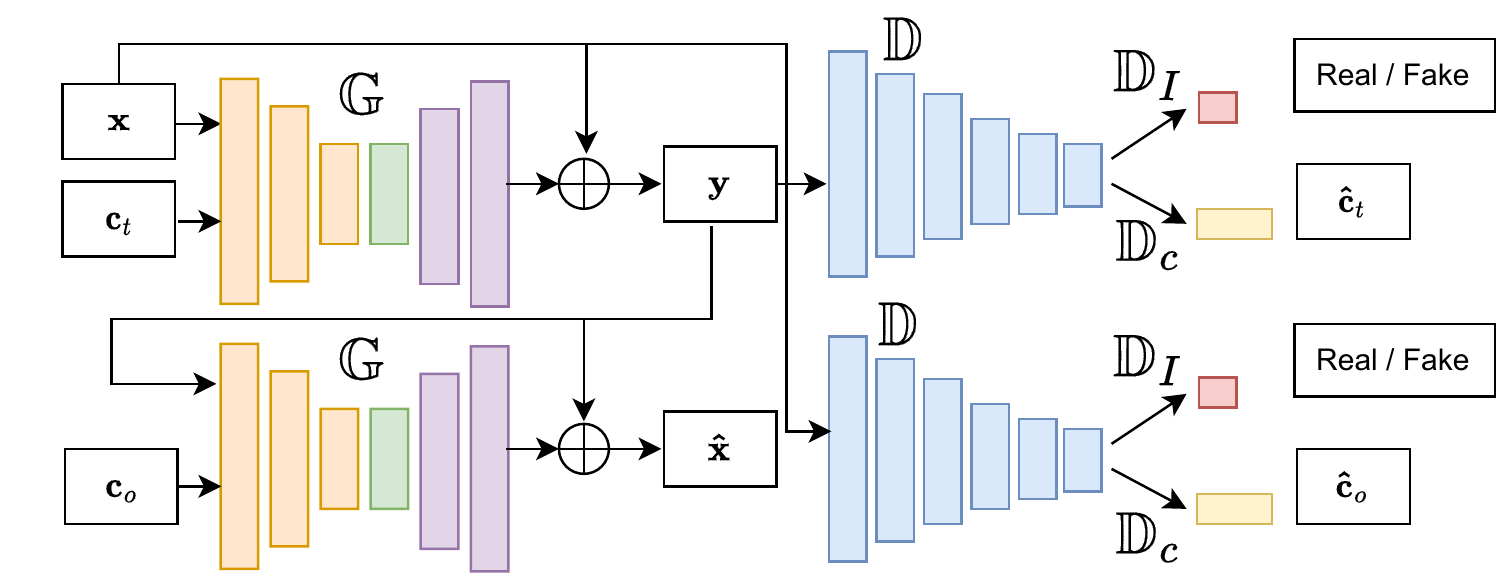}
    \caption{US-GAN consists of two modules: a generator ($\mathbb{G}$) and a discriminator ($\mathbb{D}$). $\mathbb{G}$ takes an input image ($\bm{x}$) and a target expression vector ($\bm{c}_t$) and generates an output image ($\bm{y}$) with the target expression. The same network $\mathbb{G}$ is then used to reconstruct the input image $\bm{\hat{x}}$ from $\bm{y}$ and original expression vector ($\bm{c}_o$). Discriminator head $\mathbb{D}_I$ learns to distinguish real images $\bm{x}$ from fake images $\bm{y}$ while head $\mathbb{D}_c$ classifies discriminator input into an expression class. 
    }
    \label{fig:flowchart}
\end{figure}

In this work, we aim to learn a mapping to transform an input image $\bm{x} \in \mathcal{R}^{D\times D \times 3}$ with known original expression $\mathbf{c}_o$, to an output image $\bm{y} \in \mathcal{R}^{D\times D \times 3}$ with a target expression $\mathbf{c}_t$. Both expression vectors $\mathbf{c}_o$ and $\mathbf{c}_t$ are 1-hot encodings from $C$ expression classes.
The proposed US-GAN consists of two modules: a generator and a discriminator.

\subsection{Generator}
Figure \ref{fig:flowchart} presents an overview of the complete US-GAN pipeline consisting of the following four modules. 

\subsubsection{Encoding Layers}
There are three encoding layers in our proposed network. The first layer takes an input volume $\bm{x}$ of size $D \times D \times (3+C)$ and generates a feature volume of size $D \times D \times 64$. The second encoding layer takes this feature volume, performs strided convolution for downsampling and produces an output volume of size $\frac{D}{2} \times \frac{D}{2} \times 128$. The third encoding layer takes this downsampled feature volume, performs strided convolution again and provides an encoded volume of size $\frac{D}{4} \times \frac{D}{4} \times 256$. 
    
\subsubsection{Residual Block}
Residual connections \cite{he2016deep} assist deep networks in learning better representations by directly transferring a layer's input to a later layer's output so that only the residual transformation needs to be modelled. This makes the learning task easier.
For expression synthesis, residual connections help to preserve facial details. We employ only one residual block in the body of our network. The residual block consists of two $3 \times 3$ convolutional layers of $256$ channels, two instance normalization layers \cite{ulyanov2016instance} and a ReLU layer \cite{fukushima1975cognitron}. Each convolutional layer is followed by an instance normalization layer. ReLU activation function is applied to the output of the first instance normalization layer. The input of a block is directly added after the second instance normalization layer.

\subsubsection{Decoding Layers}
The first decoding layer takes the $\frac{D}{4} \times \frac{D}{4} \times 256$ volume produced by the residual block and performs fractionally-strided convolutions to produce an upsampled volume of size $\frac{D}{2} \times \frac{D}{2} \times 128$. The second decoding layer takes these upsampled features and applies fractionally-strided convolution again to produce an upsampled volume of size $D \times D \times 64$. The last layer transforms these feature maps into an output image $\mathbb{G}(\bm{x},\mathbf{c}_t)$ of size $D \times D \times 3$.

\subsubsection{Ultimate Skip Connection}
\begin{figure}[t]
    \centering
    \includegraphics[width=.9\linewidth]{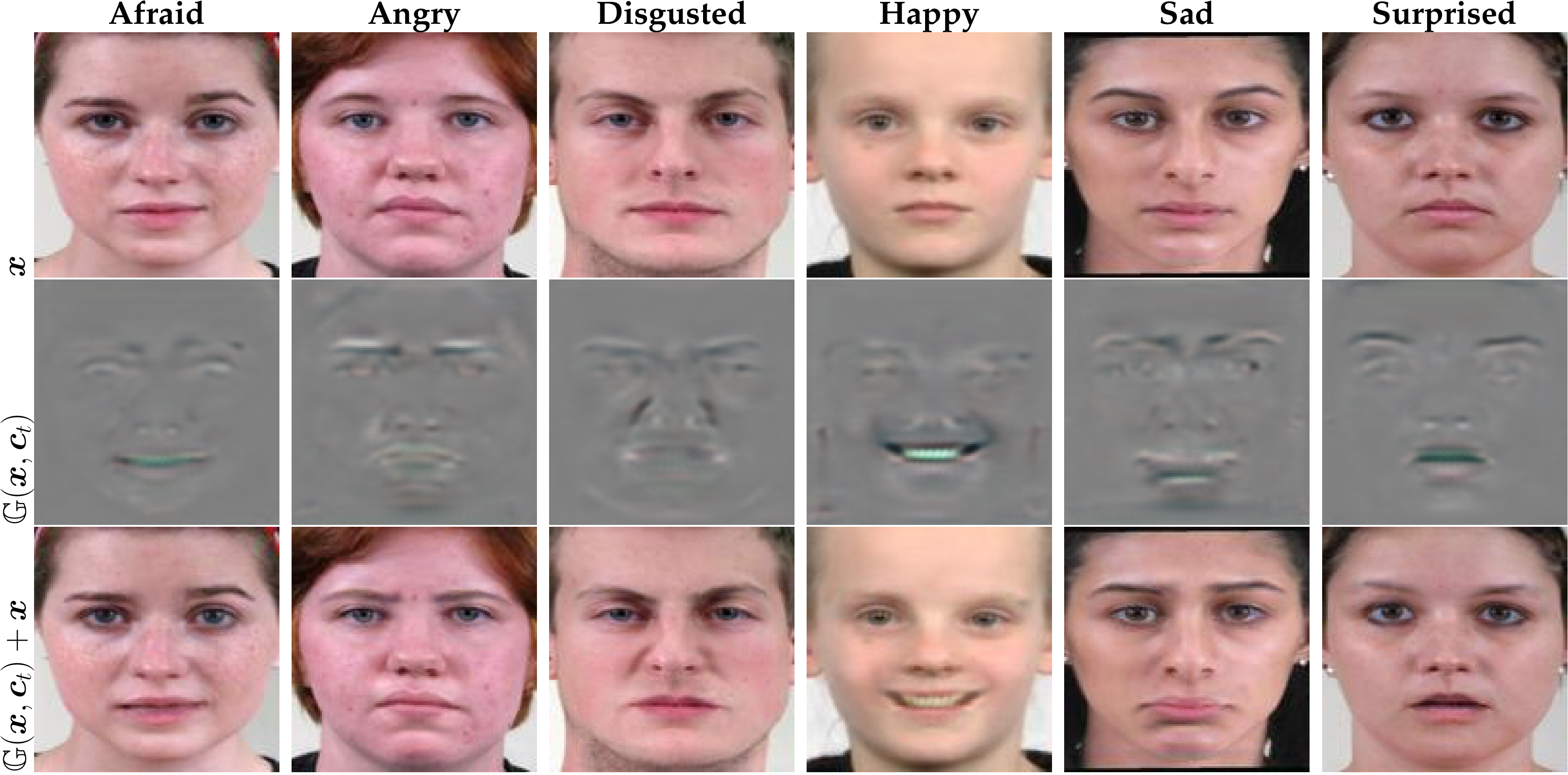}
    \caption{Illustration of residual and final output image synthesized by our proposed US-GAN. The network is encouraged to generate only expression-related details since the ultimate skip connection carries over identity, facial, and color details directly from the input image.}
    \label{fig:usgangan-results-with-residuals}
\end{figure}

For large training sets, it is possible to produce rich enough encodings and powerful enough decoders \cite{choi-2017}. However, for smaller training sets, if the input is transferred directly to the output via an ultimate skip connection, then the encoding and decoding tasks become easier. The output of the generator is obtained by adding the input to the output of the last decoding layer as
\begin{align}
     \bm{y} &=  \mathbb{G}(\bm{x}, \bm{c}_t) + \bm{x}.
\end{align}
This allows the encoding and decoding layers to focus purely on the expressions as can be seen in Figure \ref{fig:usgangan-results-with-residuals}. Since input details relating to identity, facial features, and overall color are already transferred via the ultimate skip connection, the parameters of the generator only learn to produce the residual expression $\mathbb{G}(\bm{x},\bm{c}_t)$.
Similar ideas have been explored for image restoration \cite{mao2016image} and facial attribute editing \cite{shen-2016}.

\subsection{Discriminator}
The discriminator $\mathbb{D}$ transforms its input image into a feature volume of size $\frac{D}{2^6}\times \frac{D}{2^6} \times 2048$ through a sequence of six layers of $4 \times 4$ strided convolution filters. Starting from $64$ channels in the first layer, each convolution layer doubles the the number of channels. Each convolution is followed by LeakyReLU activation. The feature volume after the sixth layer is converted into a score $\mathbb{D}_I$ that is interpreted as the chance of the discriminator's input being a real image. In parallel, the volume is also converted into a $C\times1$ vector of probabilities representing expression of the discriminator's input image.

\subsection{Loss Formulations}
\label{sec:loss_formulations}
The proposed model is trained by minimizing a combination of three loss functions. 

\subsubsection{Adversarial loss}
In order to avoid training instability and generate higher quality images, we used Wasserstein adversarial loss \cite{arjovsky2017wasserstein}, defined as 
\begin{align}
   \mathcal{L}_{A}=& E
   \left[ \mathbb{D}_I(\bm{x})\right] - E\left[\mathbb{D}_I(\bm{y}))\right] - \lambda_{gp} E \left[ (\Vert \nabla_{\bm{\bar{x}}} \mathbb{D}_I(\bm{\bar{x}}) \Vert_2 - 1)^2
    \right],
    \label{eq:adversarial_loss}
\end{align}
where $\mathbb{D}_I(\cdot)$ is proportional to the the probability of it's input image being real, $\lambda_{gp}$ is the gradient penalty coefficient, and $\bm{\bar{x}}$ is a uniform random linear combination of $\bm{x}$ and $\bm{y}$.

\subsubsection{Image Reconstruction loss}

Let $\hat{\bm{x}}$ be the reconstruction of the original image $\bm{x}$ generated from the fake expression $\bm{y}$ as
\begin{align}
    \hat{\bm{x}} &= \mathbb{G}(\bm{y},\bm{c}_o)+\bm{y}
\end{align}
In order to softly enforce that the face in the input $\bm{x}$ and the generated image $\bm{y}$ correspond to the same person, we utilize the cycle reconstruction loss \cite{zhu-2017} defined as
\begin{align}
    \mathcal{L}_{R} &= E \left[ \Vert \bm{x} - \hat{\bm{x}} \Vert_1
    \right]
\end{align}

\subsubsection{Expression Classification Loss}
We define a multiclass cross-entropy loss between target expression $\mathbf{c}_t$ and classified expression $\hat{\mathbf{c}}_t$ of the generated image $\bm{y}$ as
\begin{align}
    \mathcal{L}_{C}^{F} &= E \left[- \log \mathbb{D}_c (\bm{c}_t|\bm{y},\hat{\bm{c}}_t) \right].
    \label{eq:classification_loss_synthetic}
\end{align}
and between original expression $\mathbf{c}_o$ and classified expression $\hat{\mathbf{c}}_o$ of the input image $\bm{x}$ as
\begin{align}
    \mathcal{L}_{C}^{R} &= E \left[ - \log \mathbb{D}_c (\bm{c}_o|\bm{x}, \hat{\bm{c}}_o)\right]
    \label{eq:classification_loss_real}
\end{align}
The idea is to penalize any deviation between target and classified expression of $\bm{y}$ and between original and classified expression of $\bm{x}$.

\subsubsection{Overall GAN Objectives}
The overall objective functions for discriminator $\mathbb{D}$ and generator $\mathbb{G}$ can be written as
\begin{align}
    \mathcal{L}_{\mathbb{D}} &= - \mathcal{L}_{A} + \lambda_{C}       \mathcal{L}_{C}^{R}
    \\
    \mathcal{L}_{\mathbb{G}} &= \mathcal{L}_{A} + \lambda_{C} \mathcal{L}_{C}^{F} +  \lambda_{R} \mathcal{L}_{R}
\end{align}
where $\lambda_{C}$ and $\lambda_{R}$ denote weights of classification and reconstruction losses, respectively. Minimizing $\mathcal{L}_{\mathbb{D}}$ encourages the discriminator to improve it's ability to i) differentiate between real and fake images, and ii) classify the expression of it's input image. Minimizing $\mathcal{L}_{\mathbb{G}}$ encourages the generator to produce fake images that i) are hard to distinguish from real images, ii) have the desired expression, and iii) preserve the identity of the input image.

\subsection{Implementation Details}
\label{sub:implementation}
We train the proposed US-GAN model from scratch for $350$ epochs using the Adam optimizer \cite{kingma2014adam} with $\beta_1 = 0.5, \beta_2 = 0.999$, learning rate $0.0001$ and batch size of $8$. Following \cite{choi-2017}, we set $\lambda_C=1$, $\lambda_R=10$ and $\lambda_{gp}=10$ for all experiments.

\begin{figure}[t]
    \centering
    \includegraphics[width=.8\linewidth]{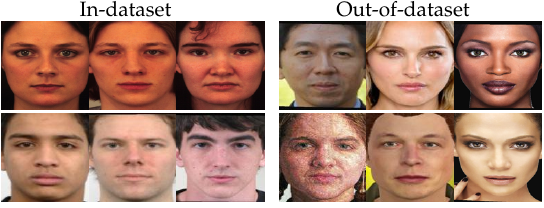}
    \caption{Example images from in- and out-of-dataset.}
    \label{fig:testing_samples}
\end{figure}

\subsection{Dataset}
\label{sub:dataset}
We trained our model using three publically available datasets KDEF \cite{KDEFDataset}, RaFD \cite{langner-2010} and CFEE \cite{du-2014}. The KDEF dataset consists of $490$ images of seven universal expressions collected from $35$ male and $35$ female participants. RaFD contains $8,040$ facial expression images collected from $67$ participants from  five different angles. We used only $469$ frontal images in our experiments. The CFEE dataset contains $5,060$ compound facial expressions images of $230$ participants. We used $1,610$ images from this dataset. In total, we use $2,569$ images from the three datasets for seven facial expressions. We used $90\%$ images for training and the rest for testing. All facial images are center-cropped and resized to $128 \times 128$. 
To evaluate the effectiveness of our proposed model on out-of-dataset images, celebrities, paintings and avatar images are downloaded from the Internet. These images are significantly different from the distribution of the training datasets. Some example images from in- and out-of-dataset are shown in Figure \ref{fig:testing_samples}.

\begin{figure}[t]
   \centering
   \includegraphics[width=.7\linewidth]{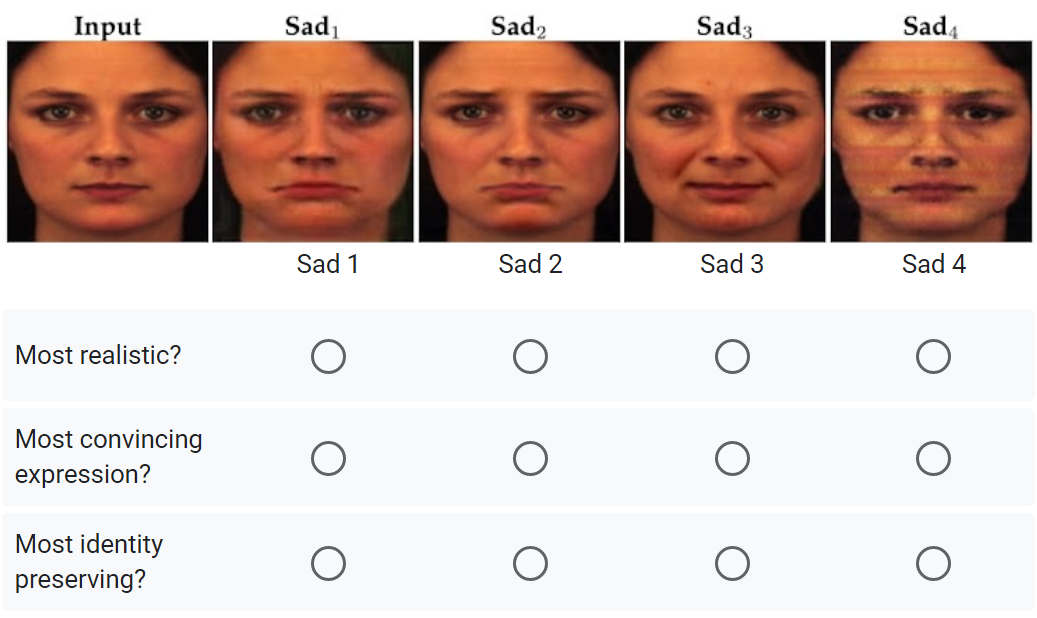}
   \caption{A sample from the user study. One input image in neutral expression along with its four manipulated versions were placed in random order in one row and human evaluators were asked to vote for the best-synthesized image in terms of i) realism, ii) mapped expression, and iii) identity preservation.}
   \label{fig:survey_examples}
\end{figure}

\subsection{Evaluation Metrics}
We compare models in terms of i) number of learnable parameters, and ii) size of training sets. We compare the quantitative performance of models in terms of the following two metrics.
\begin{enumerate}
    \item Average Content Distance (ACD) is the squared Euclidean distance between the features $\phi(\bm{x})$ of the input image and features $\phi(\bm{y})$ of the generated image. The features $\phi(\cdot)$ are extracted using a face classifier \footnote{\url{https://github.com/ageitgey/face_recognition}}. 
\begin{align}
    ACD(\bm{x},\bm{y}) = \Vert\phi(\bm{x})-\phi(\bm{y})\Vert_2^2 
\end{align}
\item Face Verification Score (FVS) computes the similarity between input and synthesized images using Face++\footnote{\url{https://www.faceplusplus.com/}} and returns a value between $0$ and $100$ to indicate the likeness between two faces.
\end{enumerate}

For qualitative comparison, we performed a user study to compute user preference percentages for StarGAN, STGAN, GANimation as well as our proposed US-GAN. Eighty users participated in this user study. Eighteen input face images from in- and out-of-dataset were randomly selected for evaluation. Synthesized expressions for these images were generated using StarGAN, STGAN, GANimation and the proposed US-GAN. One input image in neutral expression was placed on the left. It's four manipulated versions were placed next to it in random order (as shown in Figure \ref{fig:survey_examples}). The human evaluators were not made aware of the source algorithm for any manipulated image and were asked the select one image for each of the following three questions.

\begin{enumerate}
    \item Which image looks most realistic (without considering the expression)?
    \item Which image has the most convincing expression?
    \item Which image best preserves the identity of the input face?
\end{enumerate}

\begin{table}[!t]
    \centering
    \begin{tabular}{c}
       \includegraphics[width=\linewidth]{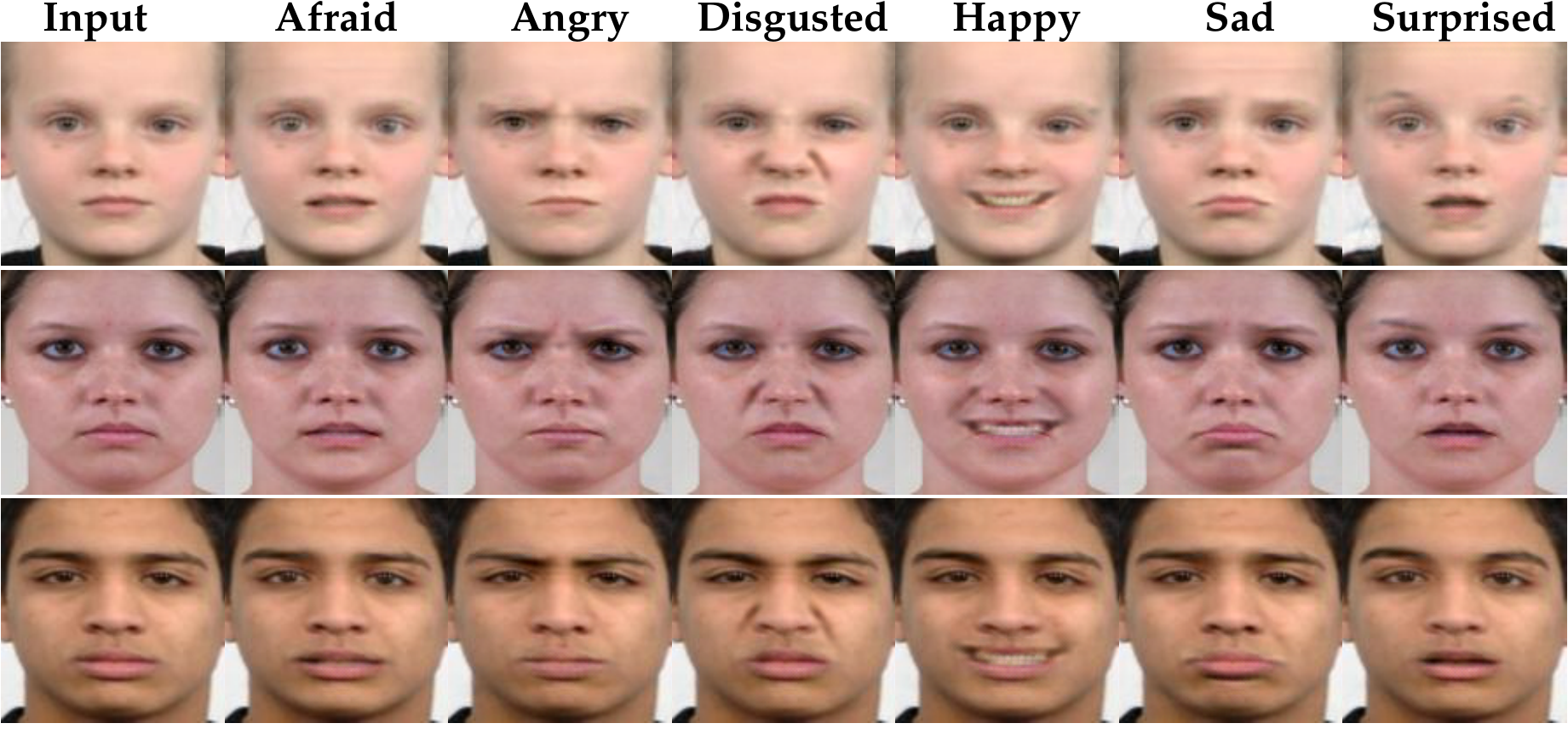}  \\
       \includegraphics[width=\linewidth]{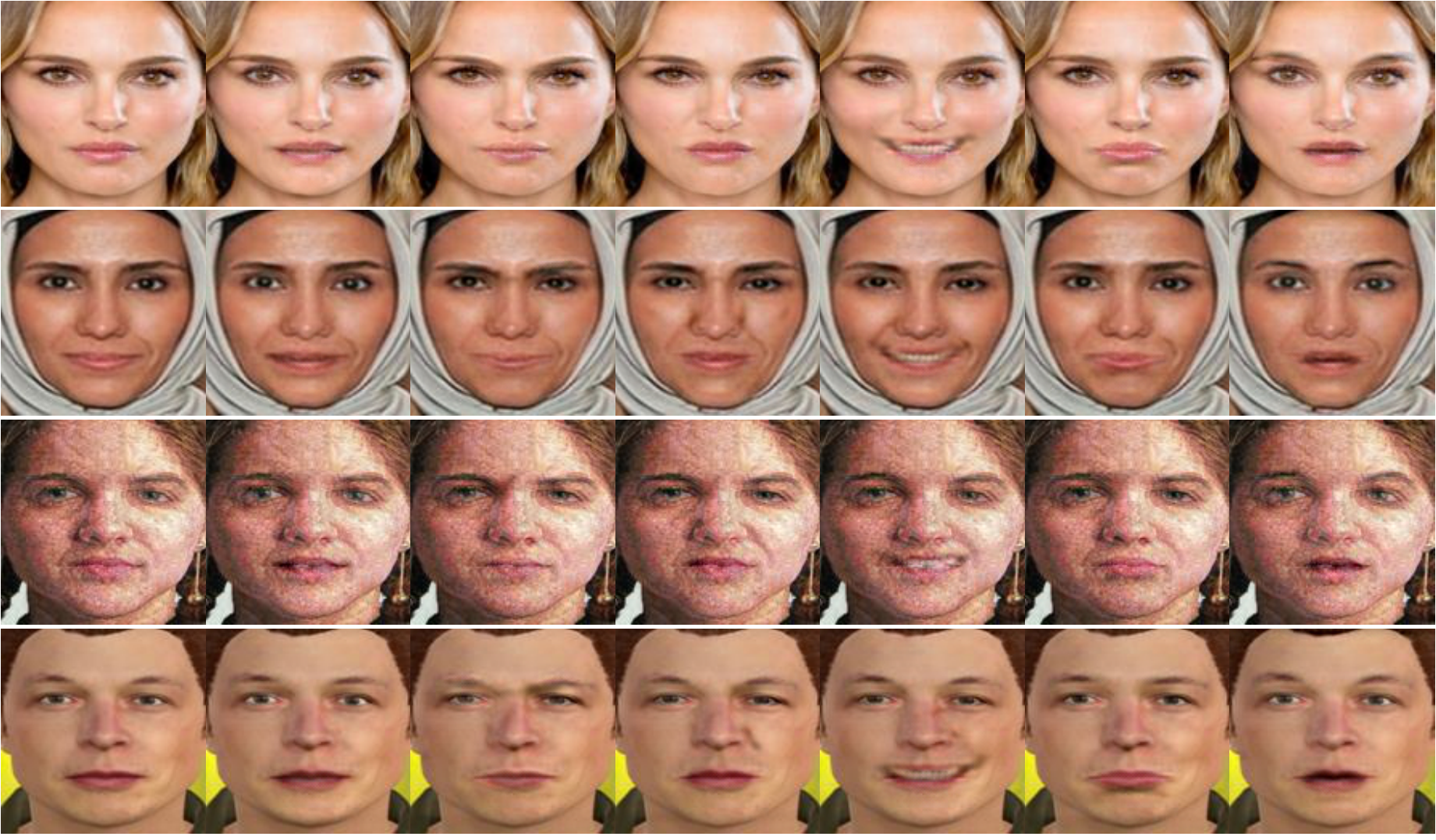}
    \end{tabular}
    \captionof{figure}{Expressions synthesised by US-GAN on \textbf{Rows 1 to 3}: in-dataset, and \textbf{Rows 4 to 7}: out-of-dataset images. Due to ultimate skip connection, the proposed method preserves input image details and colors and induces convincing expressions.
    }
    \label{fig:in_and_out-of-dataset-results}
\end{table}

\begin{figure*}[t]
    \centering
    \includegraphics[width=1\linewidth]{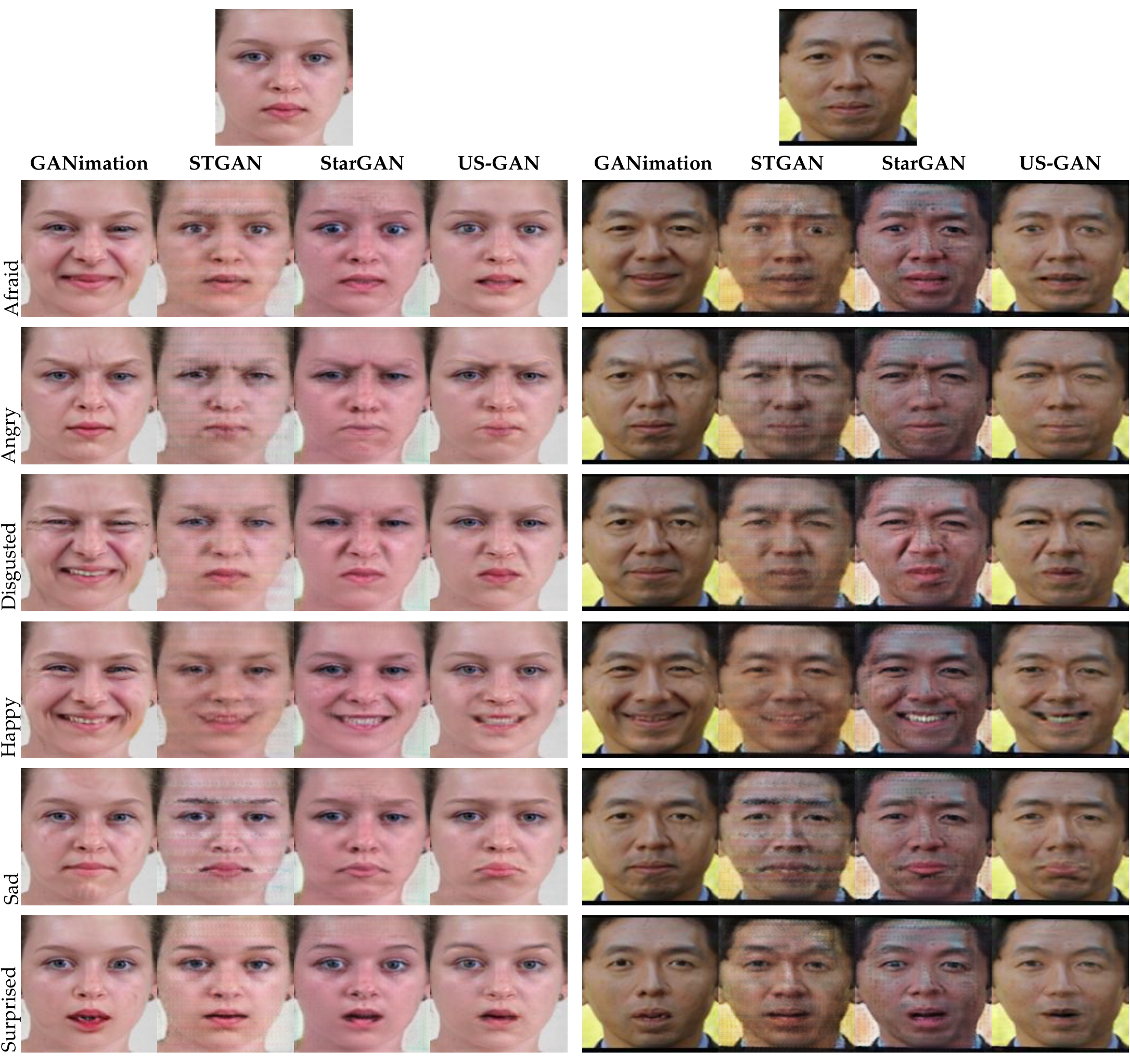}
    \caption{Comparison of facial expression synthesis results obtained by proposed US-GAN and other state-of-the-art models. \textbf{Left}: An in-dataset image. \textbf{Right}: An out-of-dataset image.} GANimation \cite{pumarola2018ganimation} introduces aging and other noticeable artifacts on all in- and out-of-dataset images. STGAN \cite{liu2019stgan} introduces pseudo-periodic artifacts. StarGAN \cite{choi-2017} introduces a pinkish bias and for out-of-dataset images, generates artifacts. The proposed US-GAN successfully synthesizes expressions while preserving identity, facial details, and color details.
    
    \label{fig:comparison_with_other_models}
\end{figure*}

\subsection{Ablation Study}
\label{sec:ablation_study}
We conducted ablation studies to investigate the effectiveness of the \textit{ultimate skip connection} and \textit{the number of residual blocks in the generator} of the proposed US-GAN. 

\subsubsection{Ultimate Skip Connection}
To qualitatively demonstrate the usefulness of the ultimate skip connection, we designed the following two experiments: 
\begin{enumerate}
    \item Train US-GAN with and without\footnote{Technically, the model should no longer be called US-GAN in this case.} ultimate skip connection to observe the difference.
    \item Train STGAN \cite{liu2019stgan} with and without ultimate skip connection to observe the difference.
\end{enumerate}

\subsubsection{Number of Residual Blocks}
In order to validate the effectiveness of residual blocks in the bottleneck of the US-GAN generator, US-GAN is trained with one and six residual blocks, respectively. We denote US-GAN model with $R$ residual blocks as US-GAN-$R$, where $R \in \{1, 6\}$. The comparison with six residual blocks is motivated by the state of the art such as StarGAN \cite{choi-2017} and GANimation \cite{pumarola2018ganimation} that uses six residual blocks.

\section{Results}
\label{sec:results}
In this section, we conduct extensive experiments to evaluate the performance of our proposed method. We first discuss baseline details in subsection \ref{sub:baseline}. Qualitative as well as quantitative evaluation on these results are presented in subsections \ref{sub:qualitative} and \ref{sub:quantitative}.

\subsection{Baselines}
\label{sub:baseline}
We compare our proposed US-GAN with three state-of-the-art, multi-domain, facial expression synthesis models, StarGAN \cite{choi-2017} , STGAN \cite{liu2019stgan} and GANimation \cite{pumarola2018ganimation}. For StarGAN and STGAN, we use the code and hyperparameter settings provided by the authors and trained on the same combined dataset (KDEF, RaFD and CFEE) used to train US-GAN. For GANimation, we used a model pre-trained on the large EmotioNet dataset \cite{fabian2016emotionet} for $30$ epochs.

\begin{figure}[t]
    \centering
    \includegraphics[width=\linewidth]{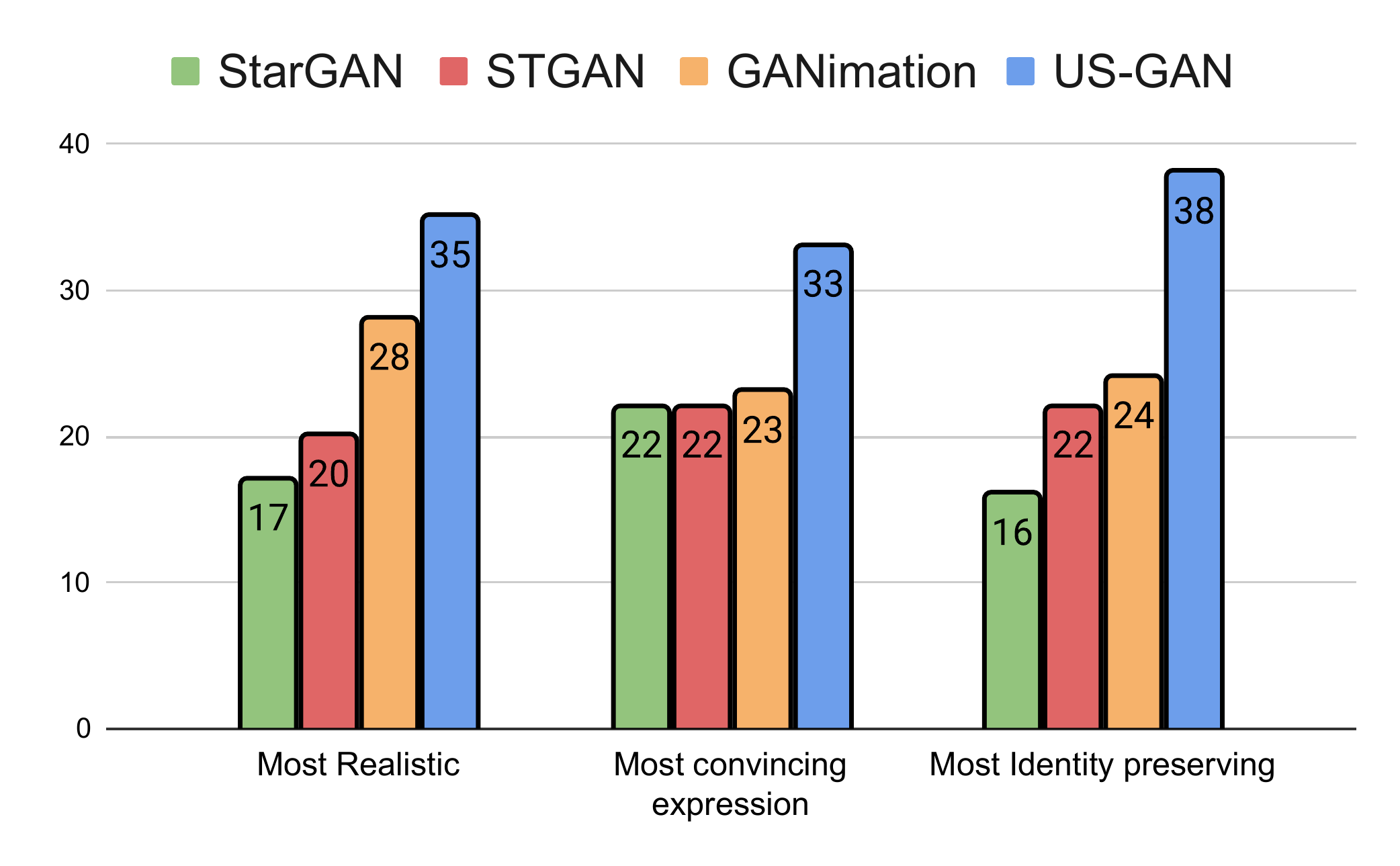}
    \caption{User study results for facial expression synthesis. The proposed US-GAN outperforms StarGAN \cite{choi-2017} , STGAN \cite{liu2019stgan} and GANimation \cite{pumarola2018ganimation} based on user preferences.}
    \label{fig:user_study_results}
\end{figure}

\subsection{Qualitative Evaluation}
\label{sub:qualitative}
Generalization of US-GAN on out-of-dataset imagery can be observed from Figures \ref{fig:main_figure} and \ref{fig:in_and_out-of-dataset-results}. The ultimate skip connection helps to transfer input image details so that the network parameters only learn to focus on generating expressions. This leads to realistic expressions and preserved identities, facial details, and color details. 
Comparison with three state-of-the-art facial manipulation models including StarGAN \cite{choi-2017}, STGAN \cite{liu2019stgan} and GANimation \cite{pumarola2018ganimation} is presented in Figure \ref{fig:comparison_with_other_models}. While existing models may induce realistic expressions on both in- and out-of-dataset images, GANimation introduces strong artifacts around the eyes, nose and mouth, StarGAN fails to recover the true input image colors, and STGAN introduces pseudo-periodic artifacts on the synthesized images. In comparison, the proposed US-GAN successfully introduces the desired expression without adding irrelevant changes.

\begin{table}[h]
\centering
   \begin{tabular}{ccccc} 
    \textbf{Method} & \textbf{\# of parameters $\downarrow$} & \textbf{\# of images $\downarrow$} & \textbf{ACD $\downarrow$} & \textbf{FVS $\uparrow$} \\ \hline
    US-GAN & 
    $\mathbf{2.5\times10^6}$ & $\mathbf{2.5\times10^3}$ & 
    \textbf{0.2832} & \textbf{94.19 $\pm$  1.11} \\
    StarGAN \cite{choi-2017} & 
    $8.5\times10^6$ & $2.1\times10^5$ & 0.5660 & 91.71 $\pm$ 1.95 \\
    STGAN \cite{liu2019stgan} & 
    $7.4\times10^7$ & $2.0\times10^5$ & 0.3725 & 92.05 $\pm$ 2.81 \\ 
    GANimation \cite{pumarola2018ganimation} & 
    $8.5\times10^6$ & $4.1\times10^5$ & 0.3899 & 87.98 $\pm$ 8.67  
    \\
    \end{tabular}
    \caption{Compared to the state of the art, the proposed US-GAN has more than three times fewer parameters and is trained on two orders of magnitude smaller dataset. It yields the best identity preservation ( lowest ACD and highest FVS) between inputs and outputs.
    }
    \label{tab:qunatitative_eval}
\end{table}

\begin{figure}[h]
    \centering
    \includegraphics[width=\linewidth]{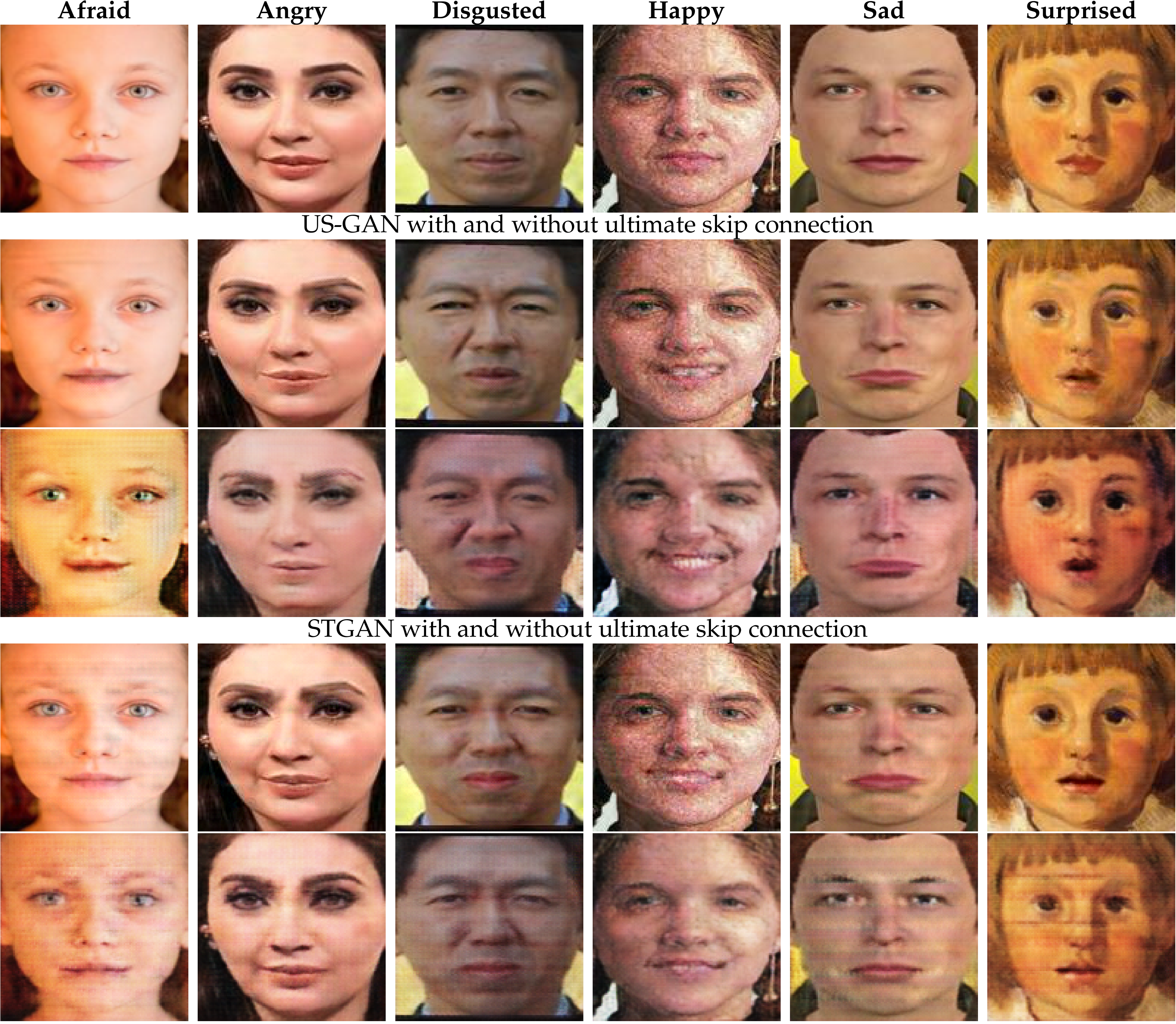}
    \caption{Impact of ultimate skip connection on generated images. \textbf{First row:} Input images. \textbf{Row 2}: US-GAN results. \textbf{Row 3}: US-GAN trained without ultimate skip connection. \textbf{Row 4}: STGAN \cite{liu2019stgan} trained after adding ultimate skip connection. \textbf{Row 5}: STGAN results. The introduction of the ultimate skip connection helps to preserve the input image as well as overall color details while introducing convincing expressions in the proposed US-GAN.In the last row, we can see that STGAN fails to preserve the color details and introduces noise-like artifacts in the synthesized images without ultimate skip connection}
    \label{fig:without-ultimate-skip}
\end{figure}

\begin{figure}[t]
    \centering
    \includegraphics[width=\linewidth]{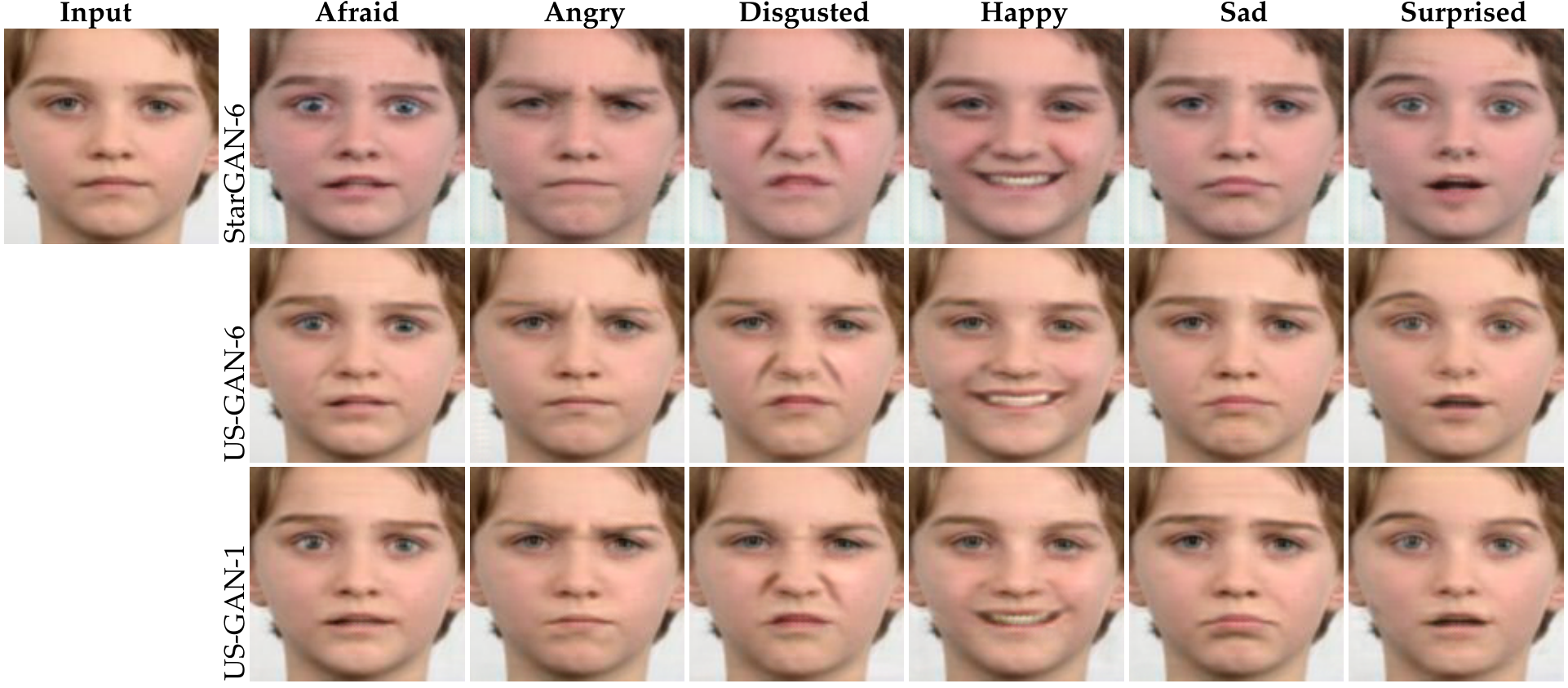}
    \caption{\textbf{First row}: StarGAN with six residual blocks failed to preserve the true colors of an input image. \textbf{Second row}: US-GAN with six residual blocks  produced sharper expressions with better preserved facial and color details. \textbf{Third row}: US-GAN with only one residual block did not suffer from significant drop in quality of results since most of the heavy lifting related to facial and color details is already carried out by the ultimate skip connection.}
    \label{fig:residual_blocks}
\end{figure}

\subsection{Quantitative Evaluation}
\label{sub:quantitative}
Table \ref{tab:qunatitative_eval} shows that our proposed method has three times fewer parameters and is trained on two orders of magnitude smaller dataset than StarGAN and GANimation. Compared to STGAN, the proposed method has an order of magnitude fewer parameters. The values of ACD and FVS indicate that US-GAN is most effective at preservation of identity and other features of the input.  

The summarized results of the user study are provided in Figure \ref{fig:user_study_results}. Compared to the second best performing model, US-GAN yielded $25\%$ improvement in realism, $43\%$ improvement in plausibility of mapped expressions, and $58\%$ improvement in identity preservation. 

\subsection{Ablation study}
\subsubsection{Ultimate Skip Connection}
We demonstrate in Figure \ref{fig:without-ultimate-skip} that the ultimate skip connection directly leads to preservation of input image details. This is true for the proposed US-GAN method as well as for STGAN \cite{liu2019stgan}. The second row contains US-GAN results. When the ultimate skip connection is removed, the third row shows that the corresponding model fails to preserve input image colors and introduces some artifacts. For STGAN as well, the fourth row shows that the addition of an ultimate skip connection leads to clear reduction in artifacts and improves the transfer of facial details from input to output.

\subsubsection{Number of Residual Blocks}
Existing models such as StarGAN \cite{choi-2017} and GANimation \cite{pumarola2018ganimation} use six residual blocks in the generator. Figure \ref{fig:residual_blocks} demonstrates that, powered by the ultimate skip connection, even one residual block can produce plausible, identity- and color-preserving transformations. In other words, the ultimate skip connection reduces the need for many parameters, which can help improve generalization. %

\section{Discussion and Future Directions}
\label{sec:discussion}
We now address a few questions raised by our results and place our results in context of existing work.

\noindent\textbf{Why an ultimate skip connection?}
Skip connections have already been used between encoding and decoding layers \cite{liu2019stgan} since they allow learning of easier residuals of intermediate tasks within deep networks. In hindsight, it seems natural to apply residual learning on the original end-to-end expression synthesis task.

The ultimate skip connection has been shown to be fundamentally important for expression synthesis since it performs the heavy lifting of transferring non-expression related details from the input to the output at no cost. This leaves the learnable parameters of the generator to focus on pure expression synthesis only.

This can be viewed from another perspective of residual learning as well. Residual learning such as the popular ResNet model \cite{he2016deep} works by learning easier, residual sub-problems within deep network layers instead of complete transformations. By incorporating a direct ultimate skip connection from input to output, we have applied residual learning to the original, end-to-end expression synthesis problem and made it easier to solve.

Since the problem has been made easier, we have solved it using fewer parameters (only one residual block in US-GAN instead of six in competing models). As a consequence, US-GAN has been shown to have better generalization on out-of-dataset imagery.

\noindent\textbf{Are residual blocks even necessary?}
Another question raised, but not answered, by our current work is whether residual blocks are necessary. This is because, even one residual block powered by an ultimate skip connection produced results better than StarGAN and GANimation models that both contain six residual blocks. Perhaps increasing direct skip connections between encoding/decoding layers in the manner of UNet \cite{ronneberger-2015} or DenseNet \cite{huang2017densely} can be more effective than residual blocks.

\noindent\textbf{What is the limitation of the proposed ultimate skip connection?}
Although the ultimate skip connection helps to recover the input image details, it slightly decreases the expressiveness of expressions as can be observed from Figure \ref{fig:without-ultimate-skip}. The ultimate skip connection helps to synthesize realistic expressions at the cost of weakened expression manipulation ability. This can perhaps be alleviated by gating the skip connection through attention mechanisms \cite{chen2020learning}.

\section{Conclusion}
\label{sec:conclusion}
We have proposed US-GAN, a smaller and more effective model for facial expression synthesis. Our primary contribution is demonstrating the benefit of an ultimate skip connection which transfers identity, facial, and color details directly from input to output. This eases the task of the generator that can then focus on inducing expressions only. It also helps to reduce the number of learnable parameters, such as multiple residual blocks, which improves generalization. Compared to state-of-the-art models, US-GAN has more than three times fewer parameters, is trained on two times smaller dataset. Based on ACD and FVS metrics, US-GAN generates realistic expressions while best preserving identity and details of the input face. US-GAN also outperforms the state of the art in terms of image and expression realism and identity preservation based on responses from human evaluations. Our results indicate that the ultimate skip connection is fundamentally important for the facial expression synthesis task. The proposed method can potentially be extended by exploring the use of intermediate skip connections as an alternative to residual blocks, and incorporating spatial attention to improve expressiveness of results.

\bibliographystyle{unsrt}  
\bibliography{references}

\begin{thebibliography}{10}

\bibitem{goodfellow-2014}
Ian Goodfellow, Jean Pouget-Abadie, Mehdi Mirza, Bing Xu, David Warde-Farley,
  Sherjil Ozair, Aaron Courville, and Yoshua Bengio.
\newblock {Generative Adversarial Nets}.
\newblock In {\em Advances in Neural Information Processing Systems}, pages
  2672--2680, 2014.

\bibitem{mirza2014conditional}
Mehdi Mirza and Simon Osindero.
\newblock {Conditional Generative Adversarial Nets}.
\newblock {\em arXiv preprint arXiv:1411.1784}, 2014.

\bibitem{perarnau-2016}
Guim Perarnau, Joost van~de Weijer, Bogdan Raducanu, and Jose~M {\'A}lvarez.
\newblock {Invertible Conditional GANs for Image Editing}.
\newblock {\em arXiv preprint arXiv:1611.06355}, 2016.

\bibitem{isola-2016}
Phillip Isola, Jun-Yan Zhu, Tinghui Zhou, and Alexei~A Efros.
\newblock {Image-to-image Translation with Conditional Adversarial Networks}.
\newblock In {\em IEEE Conference on Computer Vision and Pattern Recognition},
  pages 1125--1134, 2017.

\bibitem{zhu-2017}
Jun-Yan Zhu, Taesung Park, Phillip Isola, and Alexei~A Efros.
\newblock {Unpaired Image-to-Image Iranslation using Cycle-consistent
  Adversarial Networks}.
\newblock In {\em IEEE International Conference on Computer Vision}, pages
  2223--2232, 2017.

\bibitem{chen2020learning}
Chaofeng Chen, Dihong Gong, Hao Wang, Zhifeng Li, and Kwan-Yee~K Wong.
\newblock Learning spatial attention for face super-resolution.
\newblock {\em IEEE Transactions on Image Processing}, 30:1219--1231, 2020.

\bibitem{chen2018fsrnet}
Yu~Chen, Ying Tai, Xiaoming Liu, Chunhua Shen, and Jian Yang.
\newblock {FSRNet}: End-to-end learning face super-resolution with facial
  priors.
\newblock In {\em IEEE Conference on Computer Vision and Pattern Recognition},
  pages 2492--2501, 2018.

\bibitem{yi2017dualgan}
Zili Yi, Hao Zhang, Ping Tan, and Minglun Gong.
\newblock {DualGAN: Unsupervised Dual Learning for Image-to-Image Translation}.
\newblock In {\em IEEE International Conference on Computer Vision}, pages
  2849--2857, 2017.

\bibitem{wu2019relgan}
Po-Wei Wu, Yu-Jing Lin, Che-Han Chang, Edward~Y Chang, and Shih-Wei Liao.
\newblock {RelGAN: Multi-domain Image-to-Image Translation via Relative
  Attributes}.
\newblock In {\em IEEE International Conference on Computer Vision}, pages
  5914--5922, 2019.

\bibitem{gao2021high}
Yue Gao, Fangyun Wei, Jianmin Bao, Shuyang Gu, Dong Chen, Fang Wen, and Zhouhui
  Lian.
\newblock {High-Fidelity and Arbitrary Face Editing}.
\newblock In {\em IEEE Conference on Computer Vision and Pattern Recognition},
  pages 16115--16124, 2021.

\bibitem{choi-2017}
Yunjey Choi, Minje Choi, Munyoung Kim, Jung-Woo Ha, Sunghun Kim, and Jaegul
  Choo.
\newblock {StarGAN: Unified Generative Adversarial Networks for Multi-domain
  Image-to-Image Translation}.
\newblock In {\em IEEE Conference on Computer Vision and Pattern Recognition},
  pages 8789--8797, 2018.

\bibitem{pumarola2018ganimation}
Albert Pumarola, Antonio Agudo, Aleix~M Martinez, Alberto Sanfeliu, and
  Francesc Moreno-Noguer.
\newblock {GANimation: One-shot Anatomically Consistent Facial Animation}.
\newblock {\em International Journal of Computer Vision}, 128(3):698--713,
  2020.

\bibitem{tang2021eggan}
Junshu Tang, Zhiwen Shao, and Lizhuang Ma.
\newblock {EGGAN: Learning Latent Space for Fine-Grained Expression
  Manipulation}.
\newblock {\em IEEE MultiMedia}, 2021.

\bibitem{khan2020masked}
Nazar Khan, Arbish Akram, Arif Mahmood, Sania Ashraf, and Kashif Murtaza.
\newblock {Masked Linear Regression for Learning Local Receptive Fields for
  Facial Expression Synthesis}.
\newblock {\em International Journal of Computer Vision}, 128(5):1433--1454,
  2020.

\bibitem{akram2021-pixel_fes}
Arbish Akram and Nazar Khan.
\newblock {Pixel-based Facial Expression Synthesis}.
\newblock In {\em International Conference on Pattern Recognition}, pages
  9733--9739. IEEE, 2021.

\bibitem{chen2020domain}
Ying-Cong Chen, Xiaogang Xu, and Jiaya Jia.
\newblock {Domain Adaptive Image-to-image Translation}.
\newblock In {\em IEEE Conference on Computer Vision and Pattern Recognition},
  pages 5274--5283, 2020.

\bibitem{d2021ganmut}
Stefano d'Apolito, Danda~Pani Paudel, Zhiwu Huang, Andres Romero, and Luc
  Van~Gool.
\newblock {GANmut: Learning Interpretable Conditional Space for Gamut of
  Emotions}.
\newblock In {\em IEEE Conference on Computer Vision and Pattern Recognition},
  pages 568--577, 2021.

\bibitem{johnson2016perceptual}
Justin Johnson, Alexandre Alahi, and Li~Fei-Fei.
\newblock {Perceptual Losses for Real-time Style Transfer and
  Super-resolution}.
\newblock In {\em European Conference on Computer Vision}, pages 694--711.
  Springer, 2016.

\bibitem{ronneberger-2015}
Olaf Ronneberger, Philipp Fischer, and Thomas Brox.
\newblock {U-Net: Convolutional Networks for Biomedical Image Segmentation}.
\newblock In {\em International Conference on Medical image computing and
  computer-assisted intervention}, pages 234--241. Springer, 2015.

\bibitem{johnson-2016perceptual}
Justin Johnson, Alexandre Alahi, and Li~Fei-Fei.
\newblock {Perceptual Losses for Real-time Style Transfer and
  Super-resolution}.
\newblock In {\em European Conference on Computer Vision}, pages 694--711.
  Springer, 2016.

\bibitem{ding2018exprgan}
Hui Ding, Kumar Sricharan, and Rama Chellappa.
\newblock {ExprGAN: Facial Expression Editing with Controllable Expression
  Intensity}.
\newblock In {\em Proceedings of the AAAI Conference on Artificial
  Intelligence}, volume~32, 2018.

\bibitem{qiao2018}
Fengchun Qiao, Naiming Yao, Zirui Jiao, Zhihao Li, Hui Chen, and Hongan Wang.
\newblock {Geometry-Contrastive Generative Adversarial Network for Facial
  Expression Synthesis}.
\newblock {\em arXiv preprint arXiv:1802.01822}, 2018.

\bibitem{fabian2016emotionet}
C~Fabian Benitez-Quiroz, Ramprakash Srinivasan, and Aleix~M Martinez.
\newblock {EmotioNet: An Accurate, Real-Time Algorithm for the Automatic
  Annotation of a Million Facial Expressions in the Wild}.
\newblock In {\em IEEE Conference on Computer Vision and Pattern Recognition},
  pages 5562--5570, 2016.

\bibitem{liu2019stgan}
Ming Liu, Yukang Ding, Min Xia, Xiao Liu, Errui Ding, Wangmeng Zuo, and Shilei
  Wen.
\newblock {STGAN: A Unified Selective Transfer Network for Arbitrary Image
  Attribute Editing}.
\newblock In {\em IEEE Conference on Computer Vision and Pattern Recognition},
  pages 3673--3682, 2019.

\bibitem{wu2020cascade}
Rongliang Wu, Gongjie Zhang, Shijian Lu, and Tao Chen.
\newblock {Cascade EF-GAN: Progressive Facial Expression Editing with Local
  Focuses}.
\newblock In {\em IEEE Conference on Computer Vision and Pattern Recognition},
  pages 5021--5030, 2020.

\bibitem{mollahosseini2017affectnet}
Ali Mollahosseini, Behzad Hasani, and Mohammad~H Mahoor.
\newblock {AffectNet: A Database for Facial Expression, Valence, and Arousal
  Computing in the Wild}.
\newblock {\em IEEE Transactions on Affective Computing}, 10(1):18--31, 2017.

\bibitem{he2016deep}
Kaiming He, Xiangyu Zhang, Shaoqing Ren, and Jian Sun.
\newblock {Deep Residual Learning for Image Recognition}.
\newblock In {\em IEEE Conference on Computer Vision and Pattern Recognition},
  pages 770--778, 2016.

\bibitem{ulyanov2016instance}
Dmitry Ulyanov, Andrea Vedaldi, and Victor Lempitsky.
\newblock {Instance Normalization: The Missing Ingredient for Fast
  Stylization}.
\newblock {\em arXiv preprint arXiv:1607.08022}, 2016.

\bibitem{fukushima1975cognitron}
Kunihiko Fukushima.
\newblock Cognitron: A self-organizing multilayered neural network.
\newblock {\em Biological cybernetics}, 20(3):121--136, 1975.

\bibitem{mao2016image}
Xiaojiao Mao, Chunhua Shen, and Yu-Bin Yang.
\newblock {Image Restoration using Very Deep Convolutional Encoder-Decoder
  Networks with Symmetric Skip Connections}.
\newblock {\em {Advances in Neural Information Processing Systems}},
  29:2802--2810, 2016.

\bibitem{shen-2016}
Wei Shen and Rujie Liu.
\newblock {Learning Residual Images for Face Attribute Manipulation}.
\newblock In {\em IEEE Conference on Computer Vision and Pattern Recognition},
  pages 4030--4038, 2017.

\bibitem{arjovsky2017wasserstein}
Martin Arjovsky, Soumith Chintala, and L{\'e}on Bottou.
\newblock Wasserstein generative adversarial networks.
\newblock In {\em International conference on machine learning}, pages
  214--223. PMLR, 2017.

\bibitem{kingma2014adam}
Diederik~P Kingma and Jimmy Ba.
\newblock Adam: A method for stochastic optimization.
\newblock {\em arXiv preprint arXiv:1412.6980}, 2014.

\bibitem{KDEFDataset}
D.~Lundqvist, A.~Flykt, and A.~{\"O}hman.
\newblock {\em The Karolinska Directed Emotional Faces - KDEF, CD ROM}.
\newblock Department of Clinical Neuroscience, Psychology section, Karolinska
  Institutet, Stockholm, Sweden, 1998.

\bibitem{langner-2010}
Oliver Langner, Ron Dotsch, Gijsbert Bijlstra, Daniel~HJ Wigboldus, Skyler~T
  Hawk, and AD~Van~Knippenberg.
\newblock {Presentation and validation of the Radboud Faces Database}.
\newblock {\em Cognition and emotion}, 24(8):1377--1388, 2010.

\bibitem{du-2014}
Shichuan Du, Yong Tao, and Aleix~M Martinez.
\newblock {Compound facial expressions of emotion}.
\newblock {\em Proceedings of the National Academy of Sciences},
  111(15):E1454--E1462, 2014.

\bibitem{huang2017densely}
Gao Huang, Zhuang Liu, Laurens van~der Maaten, and Kilian~Q Weinberger.
\newblock {Densely Connected Convolutional Networks}.
\newblock In {\em Proceedings of the IEEE Conference on Computer Vision and
  Pattern Recognition}, 2017.

\end{thebibliography}

\end{document}